\documentclass{article}

\usepackage{microtype}
\usepackage{graphicx}
\usepackage{booktabs} 

\usepackage{hyperref}

\usepackage{jmlr2e}

\usepackage{url}
\usepackage[utf8]{inputenc} 
\usepackage[T1]{fontenc}    
\usepackage{hyperref}       
\usepackage{booktabs}       
\usepackage{amsfonts}       
\usepackage{nicefrac}       
\usepackage{microtype}      
\usepackage[usenames,dvipsnames]{pstricks}
\usepackage{epsfig}
\usepackage{pst-grad} 
\usepackage{pst-plot} 
\usepackage{amsmath}
\usepackage{amssymb}
\usepackage{theorem}
\usepackage{mathrsfs} 
\usepackage{euscript}
\usepackage{graphicx}
\usepackage{float}
\usepackage{textcomp}
\usepackage{listings}
\usepackage{xcolor}
\usepackage{cases}
\usepackage{algorithm}
\usepackage{algorithmic}
\usepackage{multirow}
\usepackage[lofdepth,lotdepth]{subfig}
\graphicspath{{figures/}}
\definecolor{labelkey}{rgb}{0,0.08,0.45}
\definecolor{refkey}{rgb}{0,0.6,0.0}
\definecolor{Brown}{rgb}{0.45,0.0,0.05}
\definecolor{dgreen}{rgb}{0.00,0.49,0.00}
\definecolor{dblue}{rgb}{0,0.08,0.75}
\definecolor{nido}{rgb}{0.6,0.0,0.4}
\renewcommand{\leq}{\ensuremath{\leqslant}}
\renewcommand{\geq}{\ensuremath{\geqslant}}

\newcommand{\sca}[2]{\langle {#1},{#2}\rangle}
\newtheorem{rmrk}{Remark}

\newcommand{\R}{\mathbf{R}}

\newcommand{\prox}{\mathrm{prox}}
\newcommand{\proj}{\mathrm{proj}}

\renewcommand{\epsilon}{\varepsilon}
\newcommand{\noi}{\noindent}

\newcommand{\iy}{\infty}
\renewcommand{\tilde}{\widetilde}

\begin{document}

\title{Robust supervised classification and feature selection 
using a primal-dual method}

\author{\name Michel Barlaud \\
\addr   Universit\'e C\^ote d'Azur, CNRS, I3S, France\\
\AND 
\name Antonin Chambolle \\
\addr \'Ecole Polytechnique, CNRS, CMAP, France\\
 \AND
\name Jean-Baptiste Caillau\\
\addr Universit\'e C\^ote d'Azur, CNRS, Inria, LJAD, France}

\editor{ }

\maketitle

\begin{abstract} 
This paper deals with feature selection using  supervised
classification on high dimensional datasets. A
classical approach is to project data on a low dimensional space and classify
by minimizing an appropriate quadratic cost. 
Our first contribution is to introduce a matrix of center in the definition of this quadratic cost. The benefits of are twofold: speed-up the convergence and provide a reliable signature (subset of selected genes for each class).
Moreover, as quadratic costs are not robust to outliers, we also propose to use Huber loss instead.
A classical  control
on sparsity is obtained by adding an $\ell_1$ constraint on the matrix of weights used for projecting the data.
Our second contribution is to enforce structured sparsity using a constrained formulation. To this end we propose constraints that take into account the matrix structure of the data, based either on the nuclear norm, on the $\ell_{2,1}$-norm, or on the $\ell_{1,2}$-norm for which we provide a new projection algorithm.
We optimize simultaneously the projection matrix and the matrix of centers 
thanks to a tailored constrained primal-dual method. 
 We demonstrate its effectiveness on four datasets (one synthetic, three from biological data).
Extending our primal-dual method to other criteria is easy provided that efficient
projections (on the dual ball for the loss data term, or on the constraints) are available. We establish a convergence proof of our numerical method.
\end{abstract}

\section*{Introduction}
In this paper we consider methods where feature selection is embedded into a
classification process, see \cite{Furey,Guyon}.
Sparse learning based methods have received a great attention in the last decade because of their high performance. The basic idea is to use a sparse regularizer which forces some coefficients to be zero. To achieve feature selection, the \emph{Least Absolute Shrinkage and Selection Operator} (LASSO) formulation \cite{tRS,hrtzER,ng2004,fht,Sparse,FS} adds an $\ell_1$ penalty term to the classification cost, which can be interpreted as convexifying an $\ell_0$ penalty \cite{doel,Donoho,Candes2}.
An issue is that using the Frobenius norm $\|Y\mu-XW\|_F$ (that is the $\ell_2$
norm of the vectorized matrix) for the data term is not robust to outliers.
(In the previous expression, $W$ is the projection matrix, $\mu$ the matrix of centers,
and $Y$ the binary matrix mapping each line to its class; see Section~\ref{s1}.)
Regarding structured sparsity, the most common approaches are based on {\it group LASSO} \cite{Yua,Fri,zou2006,glasso,jpvert,Vemuri,Sparse,FS} or on  $\ell_{2,1} $ norm penalty \cite{L21,Liu_L21,NIE}.
In this paper, we propose a more drastic approach that uses
an $\ell_1$ norm both on the
regularization term and on the loss function $\|Y\mu-XW\|_1 $. As a result, the
criterion is convex but not gradient Lipschitz. The idea is then to combine a splitting
method \cite{lions} with a proximal approach. Proximal methods were introduced in
\cite{Moreau} and have been intensively used in signal processing; see, \emph{e.g.},
\cite{Comb05,Silvia,Comb11,Chamb,DOS,sSN,ChaPocMAPR16}.
The first step is the computation of the proximal operator involving the affine
transform $Y\mu-XW$ in the criterion. We tackle
this point by dualizing the norm computation.
When one uses an $\ell_1$ penalization to ensure sparsity,
the computational time due to the treatment of the corresponding hyper-parameter is
expensive (see \cite{hrtzER,witten2010,myCA}). 
We propose instead a constrained approach that takes advantage of an
available efficient projection on the $\ell_1$ ball \cite{condat,duchi}.

The paper is organized as follows. We first present our setting that combines dimension
reduction, classification and feature selection. We provide in Section~\ref{s3} a
primal-dual scheme for this constrained formulation of the classification problem. In Section~\ref{s4} we lay the emphasis on structured sparsity and replace 
 the $\ell_1$  hard constraint by constraints based defined either by the nuclear norm, the $\ell_{2,1}$ norm (Group LASSO), or the $\ell_{1,2}$ norm (Exclusive LASSO). In Section~\ref{s6}, we eventually give some experimental comparisons between methods. The tests involve four different bases: a synthetic dataset, and three biological datasets (two mass-spectrometric dataset and two single cell dataset). We provide convergence proofs of our primal-dual approach in Appendix.

\section{A robust augmented variable modeling} \label{s1}
Let  $X$ be the data  $m \times d$ matrix made of $m$ line samples $x_1,\dots,x_m$
belonging to the $d$-dimensional space of features.
Let $Y \in \{0,1\}^{m\times k}$ be the label matrix where  $k \geq 2$
is the number of clusters. Each line of $Y$ has exactly one nonzero element equal
to one, $y_{ij}=1$ indicating that the sample $x_i$ belongs to the $j$-th cluster.
Projecting the data in lower dimension is crucial to be able to separate them accurately.
Let $W\in \mathbb{R}^{d\times k}$ be the projection matrix,
where $k \ll d$. Note that the dimension of the projection space is equal to the number of clusters.
The classical approach is to minimize the following squared Frobenius norm (see
\cite{FS}) with a sparsity penalty:
\begin{equation}
\min_{W} \|Y-XW\|_F^2 + \lambda \|W\|_1 
\label{Fista L2}
\end{equation}
However  $X\in \mathbb{R}^{m\times d}$ and $W\in \mathbb{R}^{d\times k}$ while 
$Y \in \{0,1\}^{m\times k}$. Moreover it is well known that convergence of proximal methods solving this criterion is very slow. In order to cope with this issue, we introduce a $\mu$ matrix  $\mu\in \mathbb{R}^{k\times k}$  such that  $Y \mu  \in \mathbb{R}^{m\times k}$.\\
\begin{equation}
\min_{(W,\mu)} \|Y\mu-XW\|_F^2 + \lambda \|W\|_1 
\label{Lasso_mu}
\end{equation}
The squared Frobenius loss is smooth, thus we can use the classical Fista algorithm \cite{Fista}.
Unfortunately the Frobenius norm is not robust to outliers and one cannot decide on
the reliability of the signature, so 
we robustify the approach by
replacing the Frobenius norm by the $\ell_1$ norm of the loss term,
$\|Y\mu-XW\|_1$.
Then using the $k \times k$ matrix of centers, $\mu$, to update the loss term according to
\begin{equation}
\|Y\mu-XW\|_1=\sum_{j=1}^{k} \sum_{l \in C_j} \|(XW)(l,:)-\mu_j \|_1
\label{kmeans}
\end{equation}
where $C_j \subset \{1,\dots,m\}$ denotes the $j$-th cluster, and where $\mu_j:=\mu(j,:)$ is the $j$-th line of $\mu$. While for $\mu=I_k$ the loss is unchanged, we actually will optimize jointly in $(W,\mu)$, adding some \emph{ad hoc} penalty to break homogeneity and avoid the trivial solution $(W,\mu)=0$. 
Using both the projection $W$ and the centers $\mu$ learnt during the training set, a new query $x$
(a dimension $d$ row vector) is classified according to the following
rule: it belongs to the cluster number $j^*$ if and only if
\begin{equation} \label{eq51}
 j^* \in \arg\min_{j=1,\dots,k} \|\mu_j-xW\|_1.
\end{equation}
(In practice, there is one and only one such cluster.)
The benefit of optimizing also wrt.\ to the centers is illustrated in Section~\ref{s6}.
%

\section{Primal-dual scheme, constrained formulation} 
\label{s3}

\subsection{Classical Lagrangian formulation} 
\label{Penalty}
We propose to minimize the $\ell_1$ loss cost with an $\ell_1$ penalty term (a Lagrangian
parameter $\lambda$ is introduced) so as to promote
sparsity and induce feature selection. So, given the matrix of labels, $Y$, and the
matrix of data, $X$, we consider the following convex supervised classification
problem where both $\mu$ and $W$ are unknowns and $I_k$ the identity matrix:
\begin{equation} \label{L1}
  \min_{(W,\mu)}  \quad \|Y\mu-XW\|_1 + \lambda \|W\|_1 + \frac{\rho}{2}\|I_k-\mu\|_F^2.
\end{equation}
Note that an $\ell_2$-regularization term has been added in order to avoid the trivial
solution $(W,\mu)=(0,0)$ while maintaining the matrix of centers $\mu$ not too far away for a rank $k$ matrix spanning all directions in the low dimensional space used for projection.
(An additional hyperparameter $\rho$ is used.)
The loss is the sum of two $\ell_1$ norms, one of them containing a linear expression of the unknowns; this is an issue since there is no straightforward means to compute the corresponding prox.
A simple way to deal with this difficulty is to dualize the computation of the $\ell_1$-norm of the loss term so as to rewrite (\ref{L1}) as 
\begin{equation}
\min_{(W,\mu)} \max_{\|Z\|_\infty \leq 1} \!\! \langle Z,Y\mu-XW \rangle  +
\frac{\rho}{2}\|I_k-\mu\|_F^2  -  \delta_{B_\iy}(Z)+ \lambda \|W\|_1.  
\end{equation}

\subsection{A constrained formulation }

In this paper we consider the convex constrained supervised classification problem
\begin{equation}
\min_{(W,\mu)} \|Y \mu-XW\|_1 + \frac{\rho}{2}\|I_k-\mu\|_F^2  \quad \text{s.t.} \quad \|W\|_1 \leq  \eta,
\label{constraint}
\end{equation}
that we dualize as: 
\begin{equation}
\min_{(W,\mu)} \max_{\|Z\|_\infty \leq 1} \!\! \langle Z,Y\mu-XW \rangle  +
\frac{\rho}{2}\|I_k-\mu\|_F^2  \quad \text{s.t.} \quad \|W\|_1 \leq  \eta.
\end{equation}
A possible primal-dual / min-max algorithm is then as follows:
\[ \begin{aligned}
& W^{n+1} := \arg \min_{ W} \frac{1}{2\tau} \|W-W^n\|^2_F-\langle X^T Z^n,W\rangle \text{ s.t. } \quad   \|W\|_1\le \eta\\
&\mu^{n+1} := \arg\min_\mu \frac{1}{2\tau_\mu}\|\mu-\mu^n\|_F^2
              + \frac{\rho}{2}\|\mu-I\|_F^2
		              + \langle Y^T Z,\mu \rangle\\
& Z^{n+1} := \proj_{\{|Z_{i,j}|\le 1\}} Z + \sigma (Y(2\mu^{n+1}-\mu^n)-X (2W^{n+1}-W^n)) 
\end{aligned} \]
These proximal steps are computed as follows:
\begin{eqnarray*}
W^{n+1} &=& \arg\min_W \frac{1}{2\tau} \|W-(W^n+\tau X^TZ^n)\|^2 \text{s.t.} \quad \|W\|_1 \leq  \eta\\
&=& \proj_{\ell_1}(W^n+\tau X^TZ^n,\eta)
\end{eqnarray*}
where ($\proj_{\ell_1}(W^n+\tau X^TZ^n,\eta)$ is the projection on the $\ell_1$
ball of radius $\eta$ ).
\[ \begin{aligned}
\mu^{n+1} =\dfrac{1}{1+\tau_{\mu}\rho }(\mu^n+\rho \tau_{\mu} I -\tau_{\mu}Y^T Z^n).
\end{aligned} \]
The iteration on $Z$ is similar (and standard,
noting for instance that computing the
proximal operator of the indicatrix is the projection).
An analogous computation also allows to obtain the modification of the iteration when using the Huber function instead of the $\ell_1$-norm; this permits to soften the computation of the loss term which otherwise enforces equality of the matrices $Y\mu$ and $XW$ outside the set of sparse components.
%
 The drawback of the term
$\|Y \mu-XW\|_1$ is that it enforces equality of the two matrices out of
a sparse set, tuning the parameters to
obtain a perfect matching of the training data.
In order to soften this behaviour, we use the Huber function instead of
the $\ell_1$-norm.
Letting
$h_\delta(t) = t^2/(2\delta)$ for $|t|\le \delta$ and $|t|-\delta/2$
for $|t|\ge \delta$, we replace  $\|Y \mu-XW\|_1$ with
\begin{equation}
  h_\delta(Y\mu-XW) := \sum_{i=1}^m\sum_{j=1}^k h_\delta((Y\mu-XW)_{i,j})
\end{equation}
and consider
\begin{equation}
\min_{(W,\mu)} h_\delta(Y \mu-XW) + \frac{\rho}{2}\|I_k-\mu\|_F^2
\text{ s.t. } \quad \|W\|_1 \leq  \eta.
\label{constraintdelta}
\end{equation}
This approach ensures that, up to a sparse set of outliers,
the components of $Y\mu$ at optimality will lie at distance $\approx \delta$ of the components of $XW$.
We can  tune the primal-dual method to solve this problem, even with
acceleration. One has $h_\delta^*(s) = \delta s^2/2$ if $|s|\le 1$, $+\infty$ else, hence we find the following saddle-point problem:
\begin{equation}\label{eq:mainsaddle}
\min_{\mu,\ \|W\|_1 \leq \eta} \max_{\|Z\|_\infty \leq 1} \!\! \langle Z,Y\mu-XW \rangle  +
\frac{\rho}{2}\|I_k-\mu\|_F^2  - \frac{\delta}{2} \|Z\|^2_F.
\end{equation}
We devise the following Algorithm~\ref{algo1-proj} using a projected gradient step.
\begin{algorithm}[H]
\begin{algorithmic}[1]
\STATE \textbf{Input:} $X,Y,N,\sigma,\tau,\tau_{\mu},\eta,\delta,\rho,\mu_0,W_0,Z_0$
\FOR{$n = 1,\dots,N$}
  \STATE{$W_\text{old} := W$}
  \STATE{$\mu_\text{old} := \mu$}
  \STATE{$W := W + \tau \cdot (X^T Z) $}
  \STATE{$W :=\proj_{\ell_1}(W,\eta)$}
  \STATE{$\mu := \frac{1}{1+\tau_{\mu} \cdot \rho }(\mu_\text{old}+\rho \cdot \tau_{\mu}
  I_k-\tau_{\mu} \cdot (Y^T Z))$}
  \STATE{$ Z := \frac{1}{1+\sigma \cdot \delta} (Z + \sigma \cdot (Y(2\mu-\mu_\text{old})-X(2W-W_\text{old})))$})
  \STATE{$Z :=  \max(-1, \min(1,Z)) $)}
\ENDFOR
\STATE \textbf{Output:} $W, \mu$
\end{algorithmic}
\caption{Primal-dual algorithm, constrained case---$\proj_{\ell_1}(V,\eta)$ is the projection on the $\ell_1$
ball of radius $\eta$).}
\label{algo1-proj}
\end{algorithm}
The convergence condition (see Appendix) imposes that: 
\begin{equation}\label{condtausig}
  \sigma\left(\frac{\tau_\mu}{1+\tau_\mu(\rho/4)}
  \|Y\|^2+\tau\|X\|^2\right)<1.
\end{equation}
The norms involved in the previous expression are operator norms, that is, \emph{e.g.},
\begin{equation}
 \|X\|  = \sup_{\|W\|_{F}\le 1} \|XW\|_F
        = \sup_{\|v\|_2\le 1} \|X(:)v\|_2. 
\end{equation}
Since the problem is strongly convex with respect to variable $ \mu$, then the descent step for the corresponding variable $ \mu$ can be increased with respect to the choice in \cite{ChaPocMAPR16}.

In the particular case when $\mu=I$, the centers are fixed and one has
\begin{equation}
\min_{W} h_\delta(Y -XW) \text{ s.t. } \quad \|W\|_1 \leq  \eta.
\label{constraintdelta}
\end{equation}
The resulting saddle-point problem is
\begin{equation}
\label{eq:mainsaddlesimplified}
\min_{W,\ \|W\|_1 \leq \eta} \max_{\|Z\|_\infty \leq 1} \!\! \langle Z,Y-XW \rangle  
  - \frac{\delta}{2} \|Z\|^2_F,
\end{equation}
and we derive the following simplified algorithm: 
\begin{algorithm}[H]
\begin{algorithmic}[1]
\STATE \textbf{Input:} $X,Y,N,\eta,\delta,W_0$
\FOR{$n = 1,\dots,N$}
  \STATE{$W_\text{old} := W$}
  \STATE{$W := W + \tau \cdot (X^T Z) $}
  \STATE{$W :=\proj_{\ell_1}(W,\eta)$}
  \STATE{$ Z := \frac{1}{1+\sigma \cdot \delta} (Z + \sigma \cdot (Y -X(2W-W_\text{old})))$})
  \STATE{$Z :=  \max(-1, \min(1,Z)) $)}
\ENDFOR
\STATE \textbf{Output:} $W$
\end{algorithmic}
\caption{Primal-dual algorithm, constrained case---with $\mu=I$ }
\label{algo1-mu=I}
\end{algorithm}
\noindent The convergence condition imposes that 
\begin{equation}\label{eq:condtausig}
 \tau \sigma\left(\|X\|^2\right)<1.
\end{equation}
The main advantage of this simplified algorithm is the reduced number of parameters to be tuned. We will compare the accuracy of the two approaches in the numerical experiments.

\section{Structured sparsity }
\label{s4}
Although results on the problem $\ref{L1}$ are available using proximal methods, little work on projections on structured constraints projections is available. This section deals with the following structured constraint sparsity methods: nuclear constraint, Group LASSO and Exclusive LASSO methods.

\subsection{Projection on the  nuclear norm }
In applications, it is often important not to forget the matrix structure of the projection matrix $W$. To preserve this information, instead of the $\ell_1$-norm one can consider the nuclear norm $\|W\|_*$, that is the sum of the singular values of $W$. Note that the nuclear norm is very popular for matrix completion \cite{Nuclear}.
The projection on the nuclear ball of of radius $\eta_{\star}$ can be computed according to Algorithm~\ref{algo1-nuclear}.
\begin{algorithm}[H]
\begin{algorithmic}[1]
\STATE \textbf{Input:} $V \eta_{\star}$
  \STATE($U,\Sigma, V)=SVD(W)$
  \STATE{$\Sigma_{\star} :=\proj_{\ell_1}(\Sigma,\eta_{\star})$}
  \STATE{$W:=U \Sigma_{\star} V$}
\STATE \textbf{Output:} $W$
\end{algorithmic}
\caption{Projection on the nuclear ball of of radius $\eta_{\star}$.---$\proj_{\ell_1}(V,\eta_{\star})$ is the projection on the $\ell_1$}
\label{algo1-nuclear}
\end{algorithm}
\noindent The complexity of computing the SVD of $W$ is $O( d \times k^2 + k^3)$. Although $d$ is large, the number of classes $k$ is small, so the algorithm is scalable (see Table~\ref{tableThyroid}).

\subsection{Projection on the $\ell_{2,1} $ norm (Group LASSO)}
The Group LASSO was first introduced in \cite{Yua}. The main idea of Group LASSO is to enforce models parameters for different classes to share features. Group sparsity reduce complexity by eliminating entire features.
Group LASSO consists in using the $\ell_{2,1} $ norm for the constraint on $W$.
The row-wise $\ell_{2,1}$ norm of a $d \times k$ matrix $W$ (whose rows are denoted $w_i$, $i=1,d$) is defined as follows:
\[ \|W\|_{2,1} := \sum_{i=1}^d \|w_i\|. \]
We use the standard following approach to compute the projection $W$ of a $d \times k$ matrix $V$ (whose rows are denoted $v_i$, $i=1,d$) on the $\ell_{2,1}$-ball of radius $\eta$:
compute $t_i$ which is the projection of the vector $(\|v_i\|_i)_{i=1}^d$ on the $\ell_1$ ball of $\R^n$ of radius $\eta$;
then, each row of the projection is obtained according to
\[ w_i = \frac{t_i v_i}{\max\{t_i,\|v_i\|\}}\,,\quad i=1,\dots,d. \]
This last operation is denoted as $W := \proj_{\ell_2}(V,t)$ in Algorithm~\ref{algo1-proj_L21}. 

\begin{algorithm}[H]
\begin{algorithmic}[1]
\STATE \textbf{Input:} $V,\eta$
  \STATE{$t :=\proj_{\ell_1}((\|v_i\|_i)_{i=1}^d,\eta)$}
  \STATE{$W := \proj_{\ell_2}(V,t)$}
\STATE \textbf{Output:} $W$
\end{algorithmic}
\caption{Projection on the $\ell_{2,1} $ norm---$\proj_{\ell_1}(V,\eta)$ is the projection on the $\ell_1$-ball of radius $\eta$ }
\label{algo1-proj_L21}
\end{algorithm}

\noindent This  algorithm requires the projection projection of the vector $(\|v_i\|_i)_{i=1}^d$ on the $\ell_1$ ball of $\R^n$ of radius $\eta$ whose complexity is only $O (d \times \log(d))$ (see Table~\ref{tableThyroid}).
Note than another approach was proposed in \cite{Liu}. The main drawback of their method is to compute the roots of an equation using bisection, which is quite slow. 

\subsection{Projection on the $\ell_{1,2} $ norm (Exclusive LASSO) }
Exclusive sparsity or exclusive LASSO was first introduced in \cite{zhou10}. The main idea of Exclusive LASSO is to enforce models parameters for different classes to compete for features.
It means that if one feature in a class is selected (large weight), the exclusive lasso method tends to assign small weights to the other features in the same class. Given a matrix $V$, the projection on the corresponding balls consists in finding a matrix $W$ which solves:
\begin{equation}\label{eq:projl12}
\min_{W} \sum_{i,j}|w_{i,j}-v_{i,j}|^2 \ \text{s.t.} \sum_i \left(\sum_j |w_{i,j}|\right)^2\le\eta^2.
\end{equation}
Our  approach is to introduce a Lagrange multiplier for the constraint and then compute it by a variant of
Newton's method (Algorithm~\ref{algoL1,2}, see details in Appendix~\ref{app:l12}).

\begin{algorithm}[H]
\begin{algorithmic}[1]
\STATE \textbf{Input:} $V, \eta$
\STATE{Sort  in decreasing order $|v_{i,j}|$ \quad for all i }
\STATE{$S_{i,p_i}:=\sum_{j=1}^{p_i}|v_{i,j}|$ \quad for all i}
\STATE{$\lambda^0 = \max_{p \
in \{1,\dots,m\}} \frac{\frac{1}{\eta}\sqrt{\sum_i  S_{i,p}^2}-1}{p}$}
\STATE{$p^0_i = \arg\max_{p_i\in\{1,\dots,m\}} S_{i,p_i}/(1+\lambda^0 p_i)$}
\STATE{$ if \quad \sum_{i=1}^n  \left(\frac{S_{i,p_i^o}}{1+\lambda^o p_i^o}\right)^2 \le \eta^2 ,\textbf{terminate} $}
\FOR{$k = 1,\dots,K$}
\STATE{$\lambda := \lambda + \frac{   \sum_{i=1}^n  \left(\frac{S_{i,p_i}}{1+\lambda p_i}\right)^2 - \eta^2 }{
  2\sum_{i=1}^n p_i \frac{\Big(S_{i,p_i}\Big)^2}{(1+\lambda p_i)^3}}$}
\FOR{$i = 1,\dots,n$}
\STATE{$p_i:=\arg\max_{p_i\in\{1,\dots,m\}} \frac{S_{i,p_i}}{1+\lambda p_i}.$}
 \ENDFOR
 \ENDFOR
\STATE{$ \delta_i = \lambda^k\frac{S_{i,p_i}}{1+\lambda {p_i}}$}
\STATE \textbf{Output: $w_{i,j} = (|v_{i,j}|-\delta_i)^+\textup{sgn}v_{i,j}$ }
\end{algorithmic}
\caption{Projection on the $\ell_{1,2} $ ball.}
\label{algoL1,2}
\end{algorithm}

\noindent The main cost in this comptutation is the sum on the rows to update $\lambda$,
\[
\sum_{i=1}^n  \left(\frac{S_{i,p_i}}{1+\lambda p_i}\right)^2.
\]
Note that as iterations progress, the matrix $S$ becomes sparse with only $n'\le n$ nonzero rows, so 
the cost decreases rapidly.
We use the brute force to compute
\[
p_i:=\arg\max_{p_i\in\{1,\dots,m\}} \frac{S_{i,p_i}}{1+\lambda p_i},
\]
noting that $m$ is small, and that the computation is stopped as soon as the maximum is reached.


\section{Numerical experiments} \label{s6}
\subsection{Experimental settings}
Our primal-dual method can be applied to any classification problem with feature
selection on high dimensional dataset stemming from computational biology, image
recognition, social networks analysis, customer relationship management, \emph{etc.}
We provide an experimental evaluation in computational biology on simulated and real
single-cell sequencing dataset. There are two advantages of working with such biological
datasets. First, many public data are now available for testing reproductibility;
besides, these datasets suffer from outliers ("dropouts") with different levels of noise
depending on sequencing experiments.
Single-cell is a new technology which has been
elected "method of the year" in 2013 by {\em Nature
Methods} \cite{SingleCellMthdYear}. 
We provide also an evaluation on proteomic and metabolic mass-spectrometric dataset.
Feature selection is based on the sparsity inducing $\ell_1$ constraint.
The projection on the  $ \ell_1 $ ball $Proj(V,\eta)$ aims at sparsifying the $W$ matrix. In class $k$, the gene $j$ will be selected if $|W(j,k)|> \epsilon $. The set of non-zero column coefficients is interpreted as the signature of the corresponding class.\\
We use the Condat method \cite{condat} to compute the projection on the  $\ell_1$-ball.
We report the classical accuracy versus $\eta$ using four folds cross validation.
Processing times are obtained on a laptop computer using an i7 processor (3.1 Ghz).
In our experiments, we normalize the features according to $\|X\|=1$, and
we set $\mu^0=I_k$, $\delta=1$ and $\rho=1$. We choose $\eta$ in connection with the desired number of genes. As $\eta$ are bounded, we can set  for $\tau =1$.  $\tau_\mu= 
\frac{\beta}{2\sqrt{m}\|Y\|-(1/4)\beta \rho}$. Then we tune $\beta$ and compute $\sigma$ using equation (\ref{condtausig}).

\subsection{Datasets}
\textbf{Simulated dataset.} We build a realistic simulation of single cell sequencing experiments. The dataset 
is composed of 
$600$ samples 15,000 genes and $k=4$ clusters.

\textbf{Dataset: Ovarian \cite{FEFA}.} The data available on UCI data base were obtained from two sources: the National Cancer Institute (NCI) and the Eastern Virginia Medical School (EVMS). All the data consist of mass-spectra obtained with the SELDI technique. The samples include patients with cancer (ovarian or prostate cancer), and healthy or control patients. The dataset is composed of $216$ samples and $15000$ features.\\ 

\noindent \textbf{Dataset: Thyroid Metabolic dataset.} The data were collected at the  University Hospital Centre. All the data consist of mass-spectra. The dataset is composed of 25 patients with cancer and 25 healthy or control patients and  $979$ features.\\

\noindent \textbf{Single cell scRNA-seq dataset.}
Zeisel et al. (\cite{zeisel}) collected mouse cells from the primary somatosensory
cortex (S1) and the hippocampal CA1 region. This dataset is composed of  3,005 cells, 7,364 genes and k=7 clusters. Note that class 8 and 9 have only 20 and 60 cells respectively.\\

\noindent \textbf{Tabula Muris \cite{TMLung}.} This set is a compendium of single cell transcriptome data from the model organism Mouse musculus, containing nearly 100,000 cells from 20 organs and tissues. The data allow comparison between gene expression in cell types. Lung Tabua Muris sub-dataset is a subset of Lung organ composed of  5,400 cells, 10,516 genes and k=14 clusters. Note that class 2 has only 5 cells.

\begin{figure}[h]
\begin{center}
\includegraphics[width=0.49\linewidth,height=4.cm]{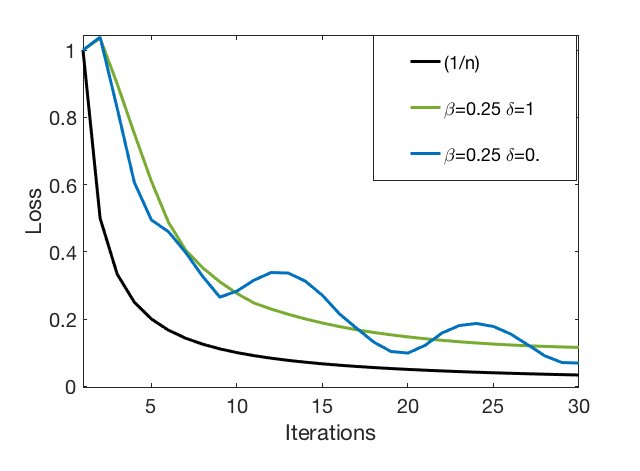}
\includegraphics[width=0.49\linewidth,height=4cm]{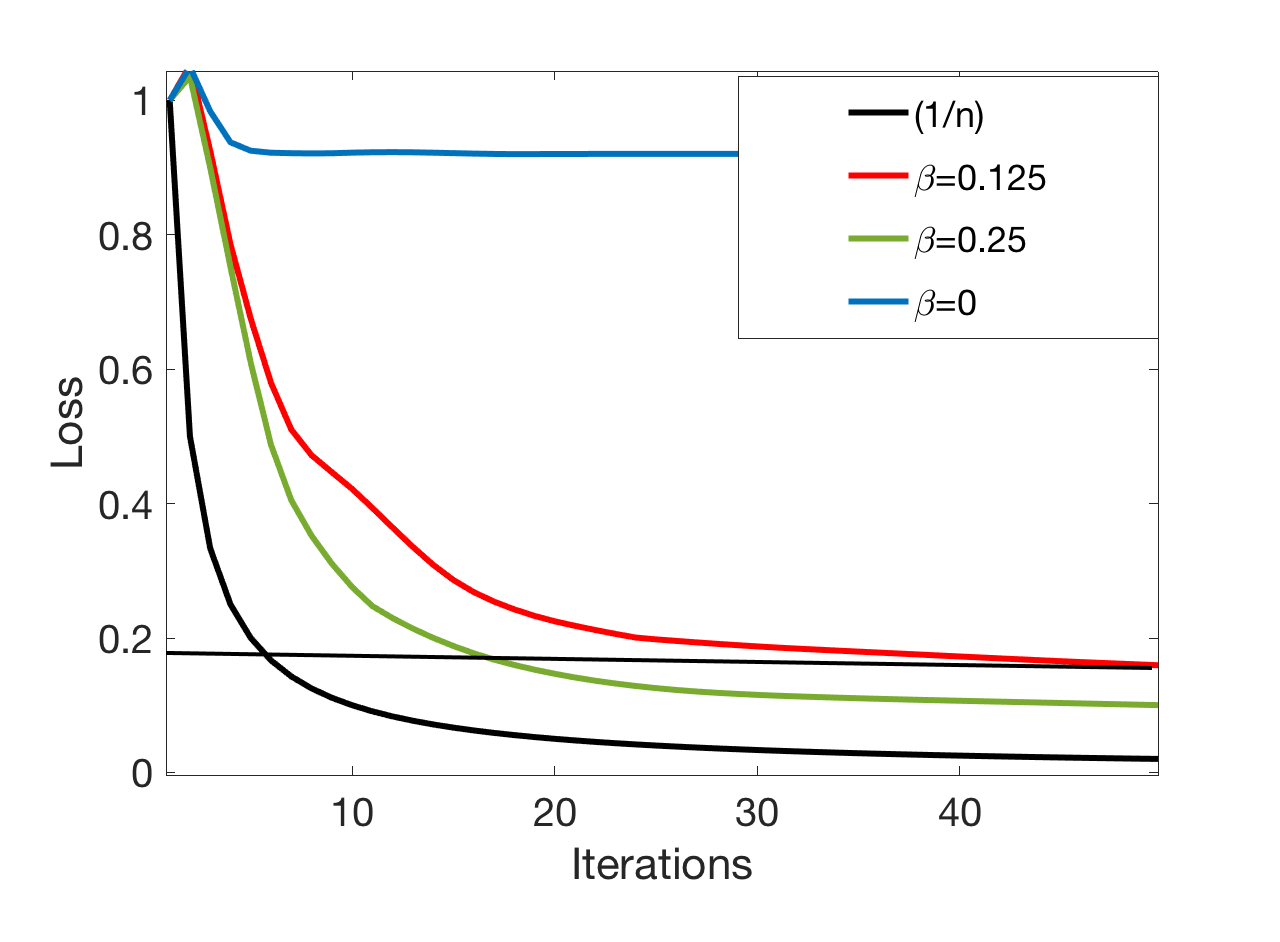}
\caption{\small Synthetic dataset.  Left
convergence of Algorithm~2 shows the benefit of using "Huber" function instead of $\ell_1$ loss (i.e. $\delta=0$). Right: This figure shows the benefit of the augmented variable $\mu$ for improving convergence. }
\label{convergence}
\end{center}
\end{figure}

\begin{figure}[h]
\begin{center}
\includegraphics[width=0.49\linewidth,height=4.cm]{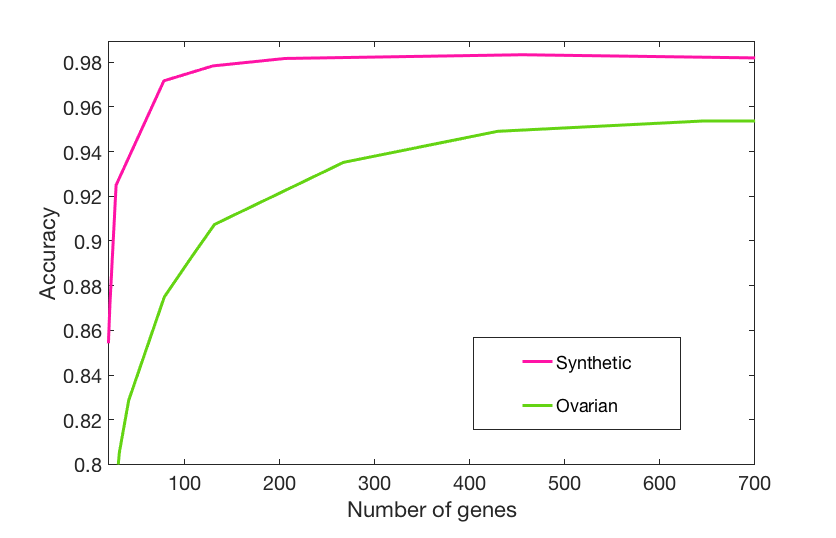}
\includegraphics[width=0.49\linewidth,height=4.cm]{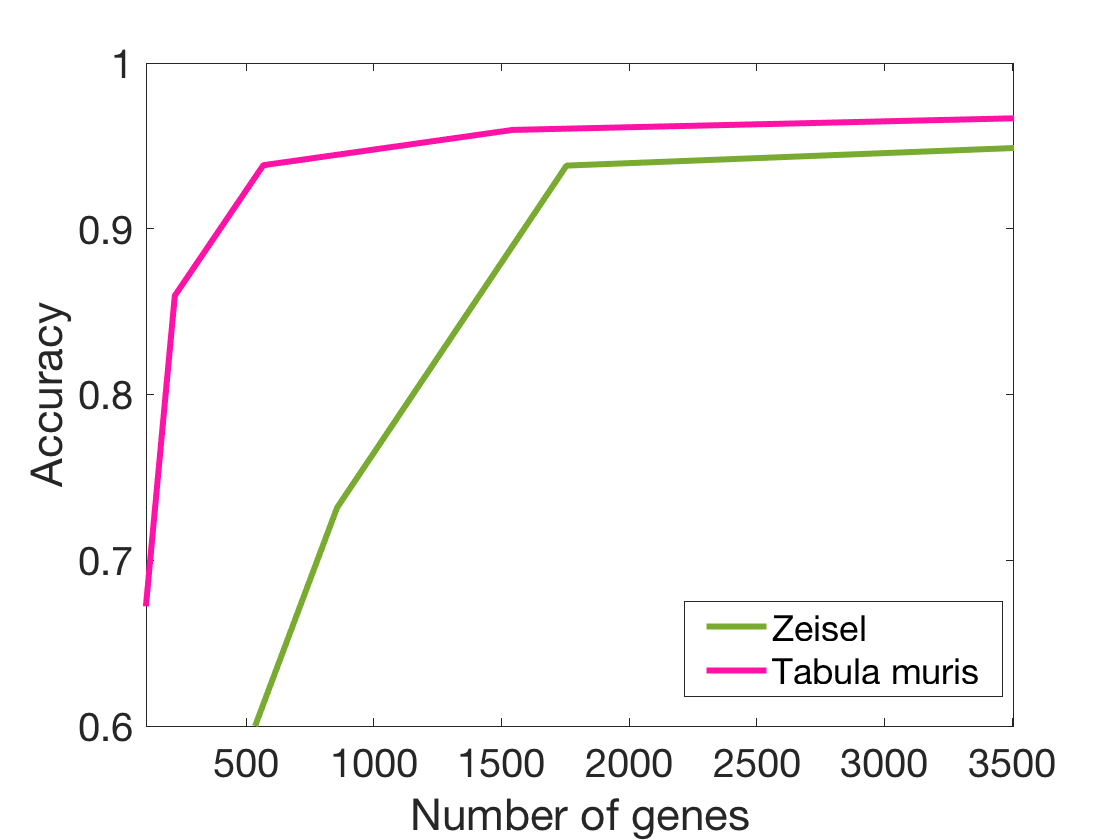}
\caption{\small Left: Synthetic and Ovarian dataset; Right Tabula Muris dataset and Zeisel dataset:  This plots show a break in
the slope of the accuracy curve versus the number of selected genes; this drastic
change can be easily detected and used to determine the relevant (and small) number of
genes to be used for the analysis.}
\label{ovarian}
\end{center}
\end{figure}

\begin{figure}[h]
\begin{center}
\includegraphics[width=0.49\linewidth,height=4.cm]{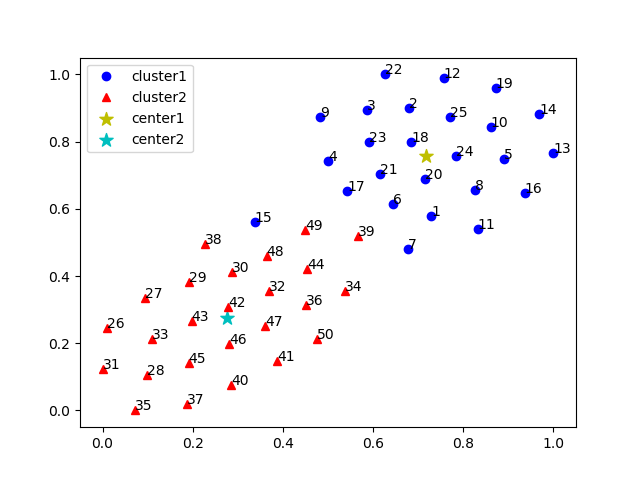}
\includegraphics[width=0.49\linewidth,height=4.cm]{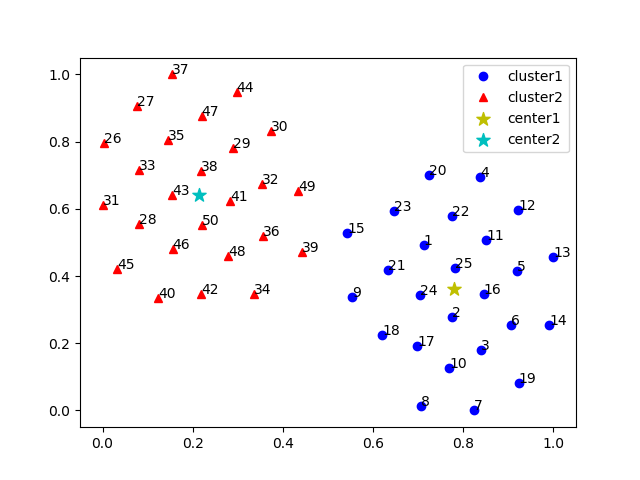}
\caption{\small Tsne \cite{tsne} of the projected data using constraint $\mu=I$ (left) and estimated $\mu$ (right)on Thyroid dataset}.
\label{Thyroid}
\end{center}
\end{figure}

\begin{table}[!htb]
    \centering
    \begin{tabular}{|c|c|c|}
        \hline
          \textbf{Methods} & Primal-dual $\mu$  & Primal-dual $\mu = I$    \\
        \hline
         Ovarian & \textcolor{red}{94.4\%} & 82.4\%  \\
         \hline
         Thyroid & \textcolor{red}{92\%} & 70\%  \\
         \hline
         Zeisel & \textcolor{red}{93.07\%} & 93.04\%  \\
         \hline
         Tabula & \textcolor{red}{97.9\%} & 97.3\%  \\
         \hline
    \end{tabular}
    \caption{Accuracy test: This table shows that accuracy using primal-dual using estimated $\mu$ outperforms primal-dual using $\mu=I$  both on Ovarian and Thyroid datasets}
    \label{tableThyroid}
\end{table}

\begin{figure}[h]	
\centering
\includegraphics[width=0.9\linewidth,height=4cm]{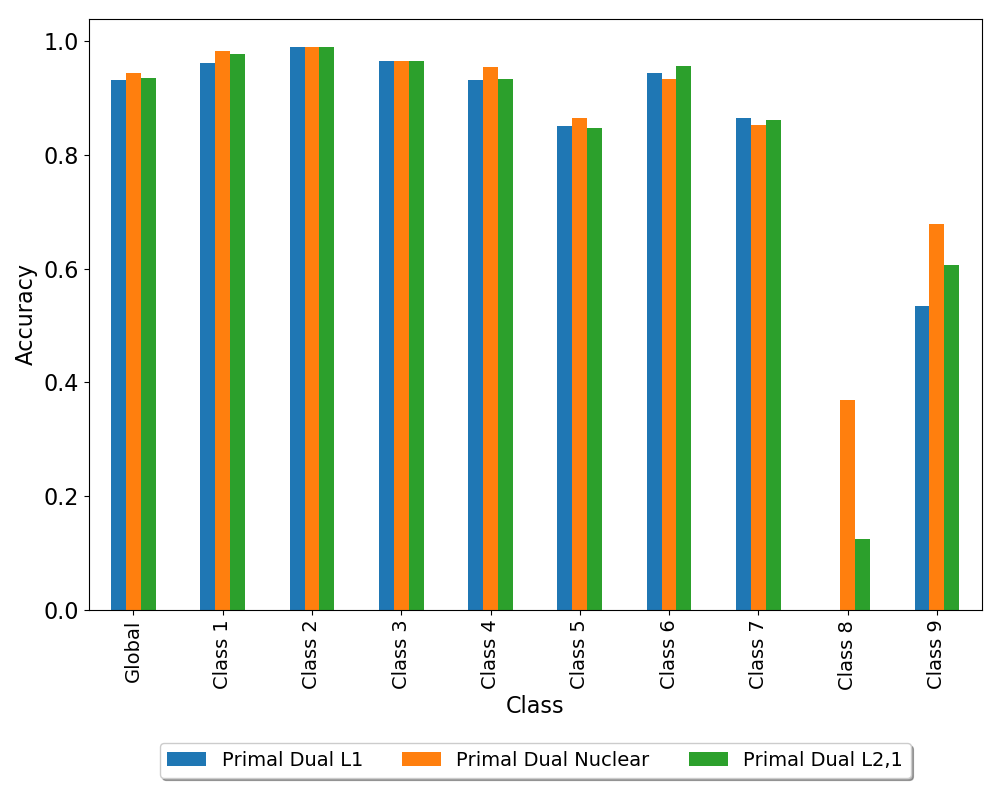}
\caption{Global accuracy and accuracy in each class: Comparison of $\ell_1 $, $\ell_{2,1} $ constraint and Nuclear constraint on Zeisel dataset  }
\label{Zeisel_bar}
\end{figure}

\begin{figure}[h]	
\centering
\includegraphics[width=0.9\linewidth,height=4cm]{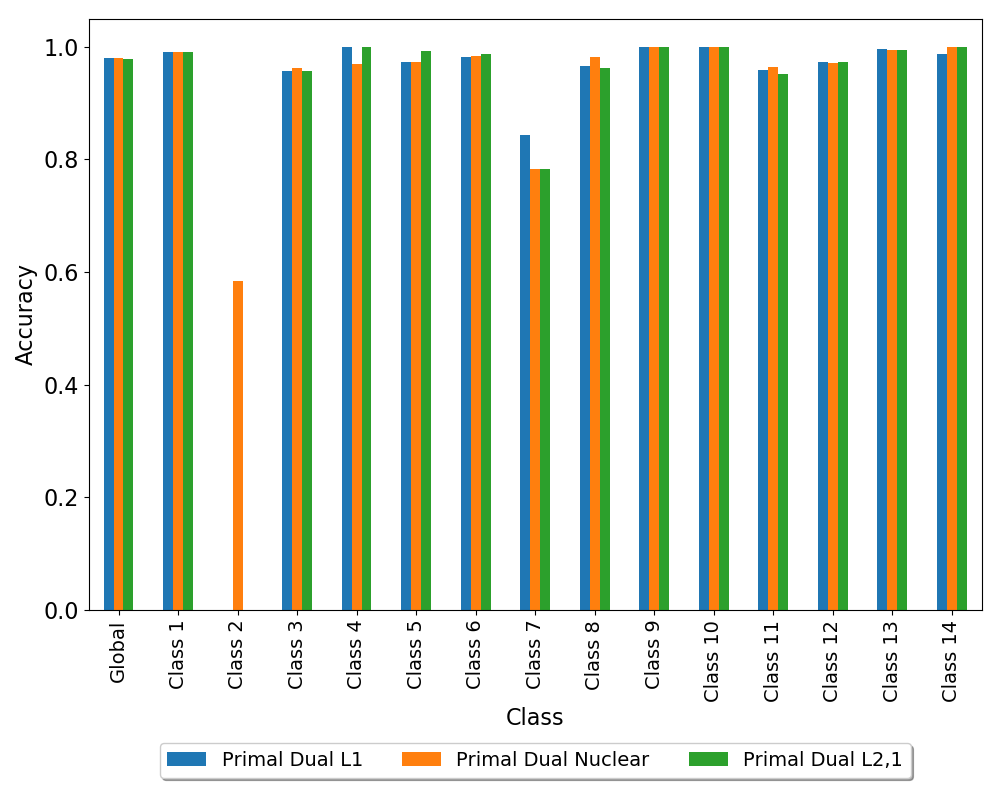}
\caption{Global accuracy and accuracy in each class: Comparison of $\ell_1 $, $\ell_{2,1} $ constraint and Nuclear constraint on Tabula Muris dataset }
\label{Tumlung_bar}
\end{figure}


\subsection{Accuracy, signature}
Figure~\ref{convergence} shows the convergence of the $\ell_1 $ loss and  Huber  loss in the training set (normalized by the value of the first iterate). Note an oscillatory convergence of the $\ell_1 $ loss while convergence of Huber loss is perfectly smooth. 
Fig.~\ref{Thyroid} and Table.~\ref{tableThyroid}, show the improvement when using adaptive $\mu$ instead of $\mu =I$ for small values of $k$.
Our primal-dual algorithm provides accuracy for each cluster. 
Fig.~\ref{Zeisel_bar} and Fig.~\ref{Tumlung_bar}, illustrating the reliability of the signature:  Nuclear norm constraint and $ \ell_{2,1}$ constraint improve accuracy small classes. Note that the standard linear regression approach does not provide accuracy.
Moreover  Fig.~\ref{ovarian}  shows a break in the slope of accuracy curve versus the number of selected genes; this drastic change of slope can be easily detected and used to determine the relevant (and small) number of genes to be used for the analysis. 

\subsection{Complexity and scalability}
Table(\ref{Time_ADMM}) shows that complexity of our primal-dual algorithm is $O(d\times m)$  for primal iterates
and $O(m\times k)$ for dual iterates. 
Note that FISTA requires that one part of the objective is smooth and the
other can be easily solved implicitly. This would be the case for
instance, for a problem of the form :
\begin{equation}
\min_{W,\mu} \|Y\mu -XW\|_F^2 + \frac{\rho}{2}\|I_k-\mu\|_F^2 +\lambda \|W\|_1 
\label{ L2}
\end{equation}
With the first squared Frobenius norm
replaced with a 1 norm, this structure is lost (also in the dual, as the
objective is strongly convex only in $\mu$) and there is no way to
implement an accelerated method (while a subgradient method would be
more expensive).
The only reasonable alternative would be ADMM, which
makes sense as long as the matrix inversions are not too hard to tackle
(here it would be very computationally expensive when $m$ and > $d$ are
large and matrix X full rank).
 Table~\ref{Time_ADMM} shows that the primal-dual method outperforms ADMM for high dimensional dataset.
\begin{table}[!h]
\begin{center}
\caption{\small Complexity of algorithms primal-dual versus ADMM (X matrix size is $m \times d,$ with m=1000). Time in milliseconds} 
\label{Time_ADMM}
\begin{tabular}{|c|c|c|c|c|c|c|c|c|}
\hline
\textbf{d } &500  &$1000$ & 2000 &  4000 & 8000 & 16000\\

\hline
Primal-dual    & 20  &60 & 170 &  397& 784  & 1620\\

\hline
ADMM   &117 & 706 & 4,630   & 32,700 &  -  &  -\\

\hline
\end{tabular}
\end{center}
\end{table}
\noindent We evaluate the complexity of the different constraint projections using random matrices of size $d \times k$ with $k=10$.
\begin{table}[!h]
\begin{center}
\caption{\small Complexity of projections ($W$  size is $d \times k,$ with $k=10$ ). Time in milliseconds} 
\label{projections}
\begin{tabular}{|c|c|c|c|c|c|c|c|}
\hline
\textbf{d } & $\ell_1$ & $\ell_{2,1}$ & Nuclear & $\ell_{1,2}$ \\
\hline
1000     &   0.75  & 0.13 & 0.46 & 1.91 \\
\hline
2000     &   1.55  & 0.51 & 0.48 & 5.1 \\

\hline
4000   & 3.12 & 0.91   & 0.89 & 12\\

\hline
8000   & 6.26 & 1.82  & 1.6 & 24\\

\hline
16000  & 13.8 & 4.01 & 3.35 &54\\

\hline
\end{tabular}
\end{center}
\end{table}
\noindent We evaluate the complexity of the different constraint projections using random matrices of size $d \times k$ with $d=1000$.
\begin{table}[!h]
\begin{center}
\caption{\small Complexity of projections ($W$  size is $d \times k,$ with $d=1000$). Time in milliseconds} 
\label{projections}
\begin{tabular}{|c|c|c|c|c|c|c|c|}
\hline
\textbf{k } & $\ell_1$ & $\ell_{2,1}$ & Nuclear & $\ell_{1,2}$ \\

\hline
10     &   0.75  & 0.13 & 0.46 & 1.91 \\

\hline
50     &   3.72  & 0.31 & 2.23 & 5.3 \\

\hline
100   & 7.8 & 0.58  & 5.01 & 10.04\\

\hline
200   & 16.3 & 1.18  & 12.3 & 19\\

\hline
500  & 47.7 & 6.44 & 59.6 &55.3\\

\hline
1000  & 99 & 16.8 & 202 & 110\\
\hline
\end{tabular}
\end{center}
\end{table}

\begin{figure}[h]
\begin{center}
\includegraphics[width=0.49\linewidth,height=5.cm]{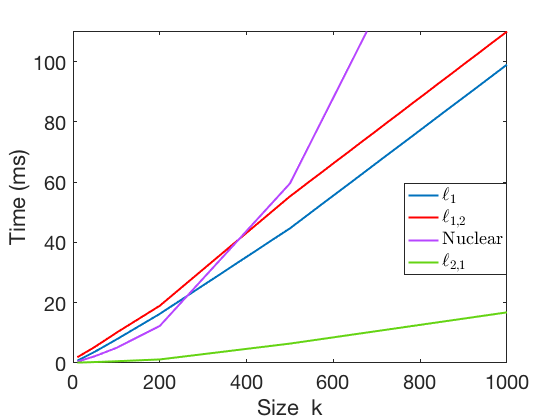}
\includegraphics[width=0.49\linewidth,height=5.cm]{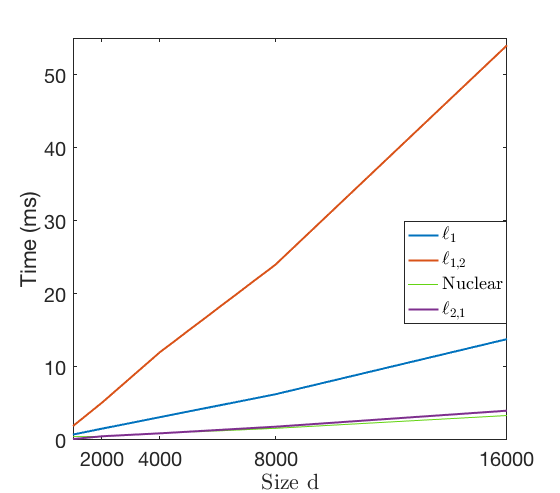}
\caption{\small This figure shows left:  time as function of size k for d =1000; right : time as function of d for k=10 for different projections.}
\label{ComparisonK}
\end{center}
\end{figure}
\begin{figure}[h]
\begin{center}
\includegraphics[width=0.49\linewidth,height=5.cm]{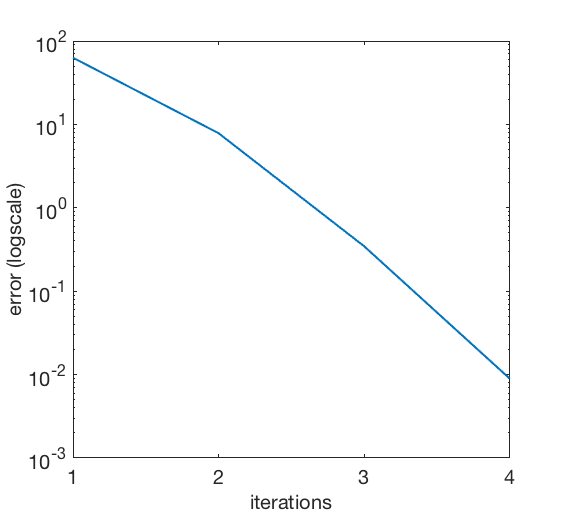}
\caption{\small This figure shows the fast convergence of Newton method (logscale).}
\label{ConvergenceNewton}
\end{center}
\end{figure}

The cost of the projection on the $ \ell_1$ ball is expected to be $O(d\times\ k\log(d\times\ k)$. The cost of the projection on the
$ \ell_{2,1}$ ball is $O(d\times\ k + d \times\ log(d))$ and thus faster than projection on the
$ \ell_1$ ball. Table.~\ref{projections} and Fig(\ref{ComparisonK})  show that for small $k$ the projection cost on the nuclear constraint is similar to projection cost on the $ \ell_{2,1}$ ball; however, for large $k$ the projection on the nuclear constraint is not scalable. Fig(\ref{ComparisonK}) shows that the cost of the projection on the
$ \ell_{1,2}$ ball is linear with d and k and slightly greater than the projection on the $ \ell_1$ ball.
Note that the complexity of the projection on the constraint (Table.~\ref{projections}) is lower than the complexity of the algorithm (Table.~\ref{Time_ADMM}).
 Thus our constrained Primal-dual method is scalable. 

\begin{table}[!h]
\begin{center}
\caption{\small Real datasets. Time in milliseconds/iteration for  Primal-dual } 
\label{tab:time}
\begin{tabular}{|c|c|c|c|c|c|c|}
\hline
\textbf{\textcolor{blue}{Dataset }} & \textcolor{blue}{Primal dual} & $\ell_1$ Projection\\

\hline
\textcolor{blue}{Synthetic m=600 , d=15,000}    & \textbf{16}    & 6.1 \\

\hline
\textcolor{blue}{Ovarian}   m=216 , d=15,000 & \textbf{6.33} &  6.1  \\

\hline
\textcolor{blue}{Zeisel}  m=3005 , d=7,364 & \textbf{36.6}  & 13.7  \\

\hline
\textcolor{blue}{Tabula} m=5400  ,  d=10,516 & \textbf{110}   & 28.8\\ 
\hline
\end{tabular}
\end{center}
\end{table}
\section{Discussion }
The goal of our paper is to provide sparse and robust features for each class. However the number of features (the sparsity) is a key issue. In order to cope with it, we propose to use accuracy in a k-fold cross validation procedure. To do so, we define $\ell_1$ centers $\mu_j$ as minimizers of the $\ell_1$ distortion in class $j$ instead of using standard centroids. With $\mu$ being the matrix of centers and Y the matrix of known labels (supervised classification), the norm of the matrix $Y\mu-XW$ is the sum of the distortions over all the classes.
The main benefits of this $\mu$ modeling is illustrated on Ovarian and  Thyroid dataset.
We emphasize two important features of our method:
(i) contrary to standard approaches based on $\ell_1$ constraints, it provides a structured signature adapted to each class;
(ii) it also provides a small number of features (sparsity) for efficient classification.
Another original point of our algorithm is that we optimize simultaneously over the centers $\mu$ and over the matrix of weights W employed for projection.
The complexity of our algorithm is linear with both variables $m $ and $d$.
Although it is not easy to carry out a fair
comparison among the different methods, due to the issue
of implementation or the choice of  parameters, we propose the following complexity comparison.
ADMM is also  reference method for tackling this problem. However, ADMM requires (large) matrix inversions and is hard to implement when these are not structured. ADMM is computationally expensive when $m$ and $d$ are large and matrix X full rank).
 Numerical experiments illustrate the benefits of our approach. 

\section{Conclusion}
We have proposed a new primal-dual method for supervised classification based on a robust Huber loss for the data and $\ell_1$, nuclear or $\ell_{2,1}$ constraints for feature selection.
Our algorithm computes jointly a projection matrix $W$ and a matrix of centers $\mu$ that are used to build a classifier. The algorithm provides a structured signature (and an estimate of its reliability) together with a minimum number of genes. We establish convergence results and show the effectiveness of our method on synthetic and biological data.
Extending the method to other criteria is  easy on condition that efficient projection (on the dual ball for the loss data term) and $\proj$ (for the regularization term) algorithms are available.

\newpage

\appendix
\section{Regularization with constrained $\ell_{1,2} $ norm}\label{app:l12}
The problem is, given $(v_{i,j})_{i=1,\dots, n}^{j=1,\dots,m}$, to find
$w=(w_{i,j})$ which solves
\begin{equation}\label{eq:projl12}
\min_{w} \left\{ \sum_{i,j}|w_{i,j}-v_{i,j}|^2 : \sum_i \left(\sum_j |w_{i,j}|\right)^2\le\eta^2\right\}.
\end{equation}
The most direct approach is to introduce a Lagrange multiplier for the constraint and then compute it by means of
Newton's method. 
Let us therefore first consider, for $\lambda>0$:
\begin{equation}
  \min_{w} \sum_{i,j} |w_{i,j}-v_{i,j}|^2 + \lambda \sum_{i} \Big(\sum_j |w_{i,j}|\Big)^2.
\end{equation}
This has the advantage to decouple into $n$ independent minimization problems as follows:
\begin{equation}
  \sum_{i} \min_{w_{i,\cdot}} \sum_j |w_{i,j}-v_{i,j}|^2 + \lambda \Big(\sum_j |w_{i,j}|\Big)^2.
\end{equation}
We first consider the generic subproblem (dropping the index $i$):
\begin{equation}
  \min_{w_{j}} \sum_j |w_{j}-v_{j}|^2 + \lambda \Big(\sum_j |w_{j}|\Big)^2
\end{equation}
whose solution is easily seen to satisfy:
\begin{equation}
  w_{j} = \Big(|v_j|-\lambda \sum_{j'}|w_{j'}|\Big)^+ \textup{sgn} v_j.
\end{equation}
Hence, letting $\delta = \lambda\sum |w_j|$, one sees that one needs to find $\delta$ such that
\begin{equation}
  \delta = \lambda\sum_j (|v_j|-\delta)^+
\end{equation}
which has a unique solution in $[0,\max_j|v_j|]$. If $|v_j|$ are sorted in decreasing order,
one must find $p\in\{ 1,\dots,m\}$ such that if
\begin{equation}
  \delta = \frac{\lambda\sum_{j=1}^p |v_j|}{1+\lambda p}
\end{equation}
one has $|v_p|\ge \delta$, $|v_{p+1}|\le \delta$. It means in fact that
\begin{equation}
  \delta = \lambda\max_{p\in\{1,\dots,m\}}\frac{\sum_{j=1}^p |v_j|}{1+\lambda p}
\end{equation}
Indeed, one can see the previous expression as the average of $0$ with weight $1/\lambda$
and $|v_j|$, $j=1,\dots,p$: $\delta =((1/\lambda)\times 0 + \sum_{j=1}^p |v_j|)/(1/\lambda + p)$
which will increase as long as one adds terms above the average, and then decrease.
Observe in addition that
\begin{equation}
  \sum_j (|v_j|-\delta)^+ = \frac{\delta}{\lambda} = \max_{p\in\{1,\dots,m\}}\frac{\sum_{j=1}^p |v_j|}{1+\lambda p}.
\end{equation}

If we return to our original problem~\eqref{eq:projl12}, we see that one needs to find $\lambda\ge 0$ such that (assuming all $|v_{i,\cdot}|$ are sorted
in decreasing order and defining $S_{i,p}:=\sum_{j=1}^p |v_{i,j}|$):
\begin{equation}\label{eq:resolambda}
  \sum_{i=1}^n \max_{p_i\in\{1,\dots,m\}} \left(\frac{S_{i,p_i}}{1+\lambda p_i}\right)^2 = \eta^2
\end{equation}
This is found by Newton's method. The function in~\eqref{eq:resolambda} is convex (as a max of convex functions), decreasing in $\lambda$.
Starting from $\lambda^0$ and the corresponding values $p_i^0$, $i=1,\dots,n$ one should compute iteratively:
\begin{equation}
  \lambda^{k+1} = \lambda^k + \frac{   \sum_{i=1}^n  \left(\frac{S_{i,p_i^k}}{1+\lambda^k p_i^k}\right)^2 - \eta^2 }{
  2\sum_{i=1}^n p_i^k \frac{(S_{i,p_i^k})^2}{(1+\lambda^k p_i^k)^3}}
\end{equation}
and then update $p_i^{k+1}$ by finding for each $i$:
\begin{equation}
  \max_{p_i\in\{1,\dots,m\}} \frac{S_{i,p_i}}{1+\lambda^{k+1} p_i}.
\end{equation}
This process must converge as the function to invert in~\eqref{eq:resolambda} is 
convex and decreasing, in particular if $\lambda^0$ is less than the optimal
lambda it is easy to see that $(\lambda^k)$ will converge monotonically, 
increasing towards the optimal value. It is not difficult to prove
that this convergence is at least linear (with rate $1-f'(\lambda^*)/f'(\lambda^0)$ if
$f(\lambda)$ denotes the left-hand side of~\eqref{eq:resolambda} and $\lambda^*$ the solution), and it is classical that
it becomes quadratic when $\lambda^k$ is close enough to the optimum
(hence the importance of finding a good starting point).
Once this has converged, one gets the thresholds $\delta_i$ by the formula
\begin{equation}
  \delta_i = \lambda^k\frac{S_{i,p_i^k} }{1+\lambda^k {p_i^k}}
\end{equation}
and then $w_{i,j} = (|v_{i,j}|-\delta_i)^+\textup{sgn}v_{i,j}$ can be easily computed on the unsorted data.\\

\paragraph{Initial $\lambda$:} the process will converge faster is one
can find a good estimate of the optimal $\lambda$ as an initial guess.
One has for the optimal $\lambda^*$:
\[
  \max_{\vec{p}=(p_1,\dots,p_n)} \sum_i\frac{ S_{i,p_i}^2}{(1+\lambda^* p_i)^2} = \eta^2
  \ge \max_{p \in \{1,\dots,m\}} \frac{\sum_i  S_{i,p}^2}{(1+\lambda^* p)^2} .
\]
The idea here is that the max on arbitrary vectors $(p_1,\dots,p_n)$ is replaced with a (smaller) max
over vectors $(p,p,\dots,p)$ with identical coordinates. It follows easily that:
\begin{equation}\label{eq:lambdaopt}
  \lambda^* \ge \max_{p \in \{1,\dots,m\}} \frac{\frac{1}{\eta}\sqrt{\sum_i  S_{i,p}^2}-1}{p} \cdot
\end{equation}
In practice, we take the right-hand side of~\eqref{eq:lambdaopt} as initial $\lambda^0$.

\section{Extension to other criteria: Frobenius loss minimization }
\label{fro}

Our method can be extended straightforwardly to other criteria provided that  we can compute the projection on the dual ball for the loss data term.
In this paper, we study an algorithm for the Frobenius norm. Note that our approach based on a dual computation of the norm allows us to use the norm itself, instead of the squared Frobenius norm.

We consider the following criterion:
\begin{equation}
\min_{(W,\mu)} \|Y \mu-XW\|_F + \frac{\rho}{2}\|I_k-\mu\|_F^2 \text{ s.t. }
\quad \|W\|_1 \leq  \eta,
\label{L2}
\end{equation}
and dualize according to
\begin{equation}\label{L2saddle}
\min_{(W,\mu)} \max_{\|Z\|_F \leq 1} \!\! \langle Z,Y\mu-XW \rangle  +
\frac{\rho}{2}\|I_k-\mu\|^2_F \text{ s.t. } \|W\|_1 \leq  \eta.
\end{equation}

Obvious modifications of the previous scheme lead to 
Algorithm:
\begin{algorithm}[H]
\begin{algorithmic}[1]
\STATE \textbf{Input:} $X,Y,N,\sigma,\tau,\eta,\delta,\rho,\mu_0,W_0,Z_0$
\STATE{$W := W_0$}
\STATE{$\mu := \mu_0$}
\STATE{$Z := Z_0$}
\FOR{$n = 1,\dots,N$}
  \STATE{$W_\text{old} := W$}
  \STATE{$\mu_\text{old} := \mu$}
  \STATE{$W := W + \tau \cdot (X^T Z) $}
  \STATE{$W :=\proj(W,\eta)$}
  \STATE{$\mu := \frac{1}{1+\tau_{\mu} \cdot \rho }(\mu_\text{old}+\rho \cdot \tau_{\mu}
  I_k-\tau_{\mu} \cdot (Y^T Z))$}
  \STATE{$ Z :=  Z + \sigma \cdot (Y(2 \mu -\mu_\text{old})-X(2 W-W_\text{old}))$}
  \STATE{$ Z :=  Z / \max\{ 1,\| Z\|_F \})$} 
\ENDFOR
\STATE \textbf{Output:} $W, \mu$
\end{algorithmic}
\caption{Primal-dual algorithm for Frobenius loss minimization: constrained case.}
\label{algo1-proj-fro}
\end{algorithm}

\section{Accelerated Constrained primal-dual approach}
 By using strong convexity with respect to Z \cite{Chamb}, we can accelerate the dual primal algorithm as follows:
\[ \begin{aligned}
    & Z^{n+1} := \proj_{\{|Z_{ij}|\le 1\}} \left(\frac{Z + \sigma (Y\mu^n-X W^{n})}{1+\sigma\delta}\right)\\
    & \theta := \frac{1}{\sqrt{1+\delta\sigma}}, \quad \bar Z := Z^{n+1}+\theta(Z^{n+1}-Z^n);\\
    & \sigma = \sigma*\theta; \quad \tau = \frac{\tau}{\theta}; \quad \tau_\mu = \frac{\tau_\mu}{\theta};\\
    & W^{n+1} := \arg \min_{ \|W\|_1\le\eta} \frac{1}{2\tau} \|W-W^n\|^2_F-\langle X^T \bar Z,W\rangle \\
    & \mu^{n+1} :=\dfrac{1}{1+\tau_{\mu}\rho }(\mu^n+\rho \tau_{\mu} I_k -\tau_{\mu}Y^T \bar Z)\\
  \end{aligned} \]

\begin{algorithm}[H]
\begin{algorithmic}[1]
\STATE \textbf{Input:} $X,Y,N,\sigma,\tau,\eta,\delta,\rho,\mu_0,W_0,Z_0$
\STATE{$W := W_0$}
\STATE{$\mu := \mu_0$}
\STATE{$Z := Z_0$}
\FOR{$n = 1,\dots,N$}
  \STATE{$W_\text{old} := W$}
  \STATE{$\mu_\text{old} := \mu$}
  \STATE{$Z_\text{old} := Z$}
  \STATE{$W := W + \tau \cdot (X^T Z) $}
  \STATE{$W :=\proj(W,\eta)$}
  \STATE{$\mu := \frac{1}{1+\tau_{\mu} \cdot \rho }(\mu_\text{old}+\rho \cdot \tau_{\mu}
  I_k-\tau_{\mu} \cdot (Y^T Z))$}
  \STATE{$ Z := \frac{1}{1+\sigma \cdot \delta} (Z + \sigma \cdot (Y(2\mu-\mu_\text{old})-X(2W-W_\text{old})))$})
  \STATE{$Z :=  \max(-1, \min(1,Z)) $)}
  \STATE{$\theta := \frac{1}{\sqrt{1+\delta\sigma}}; \quad  Z := Z+\theta(Z- Z_\text{old});$\\
     $\sigma = \sigma*\theta; \quad \tau = \frac{\tau}{\theta}; \quad \tau_\mu = \frac{\tau_\mu}{\theta}$}\\
\ENDFOR
\STATE \textbf{Output:} $W, \mu$
\end{algorithmic}
\caption{Primal-dual algorithm, accelerated  constrained case---$\proj(V,\eta)$ is the projection on the $\ell_1$
ball of radius $\eta$ (see \cite{condat}).}
\label{algo1-proj-accel}
\end{algorithm}

An over-relaxed variant of the previous algorithm is presented below
(Algorithm~\ref{L1-SR}).
\begin{algorithm}[H] 
\begin{algorithmic}[1]
\STATE \textbf{Input:} $X,Y,N,\sigma,\tau,\tau_{\mu},\eta,\delta,\rho,\mu_0,W_0,Z_0,\gamma\in (-1,1)$
\FOR{$n = 1,\dots,N$}
  \STATE{$W_\text{old} := W$}
  \STATE{$\mu_\text{old} := \mu$}
  \STATE{$ W := W + \tau \cdot (X^T Z)$}
  \STATE{$ W := \proj( W,\eta)$}
  \STATE{$ \mu :=\frac{1}{1+\tau_{\mu} \rho }(\mu+\rho \tau_{\mu} I_k-\tau_{\mu}
  (Y^T Z))$}
   \STATE{$ Z := \frac{1}{1+\sigma \cdot \delta} \cdot (Z + \sigma \cdot (Y(2 \mu -\mu_\text{old})-X(2 W-W_\text{old})))$}
    \STATE{$Z :=  \max(-1, \min(1,Z)) $)} 
  \STATE{$W :=  W+\gamma( W-W_\text{old})$}
  \STATE{$\mu :=  \mu+\gamma ( \mu- \mu_\text{old})$}
  \STATE{$Z :=  Z + \gamma( Z-Z_\text{old})$}
\ENDFOR
\STATE \textbf{Output:} $W, \mu$
\end{algorithmic}
\caption{Primal-dual algorithm, constrained case with over-relaxation.}
\label{L1-SR}
\end{algorithm}
The convergence condition discussed in 
Appendix~\ref{s5}
imposes that 
\begin{equation}\label{eq:condbase}
     \boxed{\sigma\left(\tau_\mu\|Y\|^2+\tau\|X\|^2\right)<1},
\end{equation}

\section{Regularization with constrained elastic net }

In order to handle features with high  correlation, We consider the convex constrained supervised classification problem,
\begin{equation}
\min_{(W,\mu)} \|Y \mu-XW\|_1 + \frac{\rho}{2}\|I_k-\mu\|_F^2 + \frac{\alpha}{2}  \|W\|_F^2 
 \text{ s.t. } \quad \|W\|_1 \leq  \eta,
\label{enet}
\end{equation}
that we dualize as before: 
\begin{equation} \label{eq50}
\min_{(W,\mu)} \max_{\|Z\|_\iy \leq 1} \!\! \langle Z,Y\mu-XW \rangle  +
\frac{\rho}{2}\|I_k-\mu\|_F^2 + \frac{\alpha}{2}  \|W\|_F^2 
 \text{ s.t. } \|W\|_1 \leq  \eta.
\end{equation}
We adapt the update of $W$ of Algorithm~\ref{algo1-proj} by using
a shrinkage on $W$, and devise Algorithm~\ref{algo1-proj-elastic}.

\begin{algorithm}[H]
\begin{algorithmic}[1]
\STATE \textbf{Input:} $X,Y,N,\sigma,\tau,\eta,\delta,\alpha,\rho,\mu_0,W_0,Z_0$
\STATE{$W := W_0$}
\STATE{$\mu := \mu_0$}
\STATE{$Z := Z_0$}
\FOR{$n = 1,\dots,N$}
  \STATE{$W_\text{old} := W$}
  \STATE{$\mu_\text{old} := \mu$}
  \STATE{$W := \frac{1}{1+\tau\alpha}(W + \tau (X^T Z ))$}
  \STATE{$W :=\proj(W,\eta)$}
  \STATE{$\mu := \frac{1}{1+\tau_{\mu} \cdot \rho }(\mu_\text{old}+\rho \cdot \tau_{\mu}
  I_k-\tau_{\mu} \cdot (Y^T Z))$}
  \STATE{$ Z := \frac{1}{1+\sigma \cdot \delta} (Z + \sigma \cdot (Y(2\mu-\mu_\text{old})-X(2W-W_\text{old})))$})
  \STATE{$Z :=  \max(-1, \min(1,Z)) $\label{statefinal1}}
\ENDFOR
\STATE \textbf{Output:} $W, \mu$
\end{algorithmic}
\caption{Primal-dual algorithm, with  elastic net constrained case.}
\label{algo1-proj-elastic}
\end{algorithm}

\section{Convergence Analysis} \label{s5}

\subsection{Convergence of primal-dual algorithms} \label{s4.1}
The proof of convergence of the algorithms relies on Theorems~1 and~2 in~\cite{ChaPocMAPR16} which we slightly adapt for our setting.
The algorithms we present here correspond to Alg.~1 and~2 in that reference,
adapted to the particular case of problem~\eqref{L1}
and its saddle-point formulation~\eqref{eq:mainsaddle}. In addition, here, the
primal part of the objective is ``partially strongly convex'' ($\rho$-strongly convex with
respect to the variable $\mu$, thanks to the term $(\rho/2)\|\mu-I\|^2$. (We could exploit
this to gain ``partial acceleration''~\cite{ValPocJMIV17}, however at the expense
of a much more complex method and no clear gain for the variable $W$,
while translated in the Euclidean setting, \cite{ChaPocMAPR16} remains simple and
easy to improve.) 
For our setting we consider 
a general objective of the form:
\begin{equation}\label{eq:generalform}
\min_{x,x'}\max_y f(x)+g(x') + \sca{Kx+K'x'}{y} - h^*(y)
\end{equation}
for $f,g,h$ convex functions whose ``prox'' (see below) are easy to compute and $K,K'$ linear operators, and we assume moreover $f$ is $\rho$-strongly convex for some $\rho>0$.
We will show how this last property can 
be exploited to ``boost'' the convergence, allowing for larger steps than
usually suggested by other authors.
When computing the ``prox'' $\hat x$ at point $\bar x$ of a $\rho$-strongly convex function $x\mapsto f(x)$, with parameter $\tau$, that is, the minimizer
\begin{equation}\label{eq:prox}
\hat x =\prox_{\tau f}(\bar x) := \arg\min_x f(x)+\frac{\|x-\bar x\|^2}{2\tau},
\end{equation}
one has for all test point $x$:
\begin{equation}
f(x) + \frac{\|x-\bar x\|^2}{2\tau}\ge 
f(\hat x) + \frac{\|\hat x-\bar x\|^2}{2\tau} + \frac{\|x-\hat x\|^2}{2\tau} + \frac{\rho}{2}\|x-\hat x\|^2.
\end{equation}
However, combined with non-strongly convex iterates, the slight improvement given by the factor $\rho$
is hard to exploit (whereas for simple gradient descent type iterates one obviously can derive 
linear convergence to the optimum), see for instance~\cite{ValPocJMIV17} for
a possible strategy.
We exploit here this improvement in a different way. We combine the parallelogram identity
\[
\|x-\bar x\|^2 + \|x-\hat x\|^2 = \frac{1}{2}\|\bar x-\hat x\|^2 + 2\left\|x-\tfrac{\bar x+\hat x}{2}\right\|^2
\]
with the previous inequality to obtain:
\begin{multline}
\label{eq:improveddescent}
f(x) + (1+\tau\tfrac{\rho}{2})\frac{\|x-\bar x\|^2}{2\tau}\ge  f(\hat x) + (1+\tau\tfrac{\rho}{4})\frac{\|\hat x-\bar x\|^2}{2\tau} + (1+\tau\tfrac{\rho}{2})\frac{\|x-\hat x\|^2}{2\tau}.
\end{multline}

The first type of algorithm we consider is Algorithm~\ref{algo1}, which corresponds to
Alg.~1 in~\cite{ChaPocMAPR16} (see also~\cite{pock2009algorithm,EZCSIIMS10,Chamb}). 
It consists in tackling problem~\eqref{eq:generalform} by alternating
a proximal descent step in $x,x'$ followed by an ascent step in $y$:
\begin{equation}\label{eq:primaldual}
\begin{split}
x^{n+1} &= \prox_{\tau f}(x^n-\tau K^T y^n),\\
{x'}^{n+1} &= \prox_{\tau' g}({x'}^n-\tau' {K'}^T y^n),\\
y^{n+1} &= \prox_{\sigma h^*}(y^n +
\\ & \sigma (K(2x^{n+1}-x^n) +K'(2{x'}^{n+1}-{x'}^n))).
\end{split}
\end{equation}
We then introduce the ``ergodic'' averages
\[
X^N = \tfrac{1}{N}\sum_{n=1}^N x^n, \ 
{X'}^N = \tfrac{1}{N}\sum_{n=1}^N {x'}^n, \ 
Y^N = \tfrac{1}{N}\sum_{n=1}^N y^n.
\]
Theorem~1 in~\cite{ChaPocMAPR16}, shows with an  elementary proof the estimate, for any test point $(x,x',y)$:
\begin{equation}\label{eq:ratePD}
    \mathcal L(X^N,{X'}^N,y) - \mathcal L(x,x',Y^N) \le 
    \frac{1}{2N} \left\|\begin{pmatrix} x\\x'\\ y\end{pmatrix} - \begin{pmatrix} x^0\\{x'}^0\\ y^0\end{pmatrix}\right\|^2_{M_{\tau,\tau',\sigma}}
\end{equation}
where $\mathcal L$ is the Lagrangian function in~\eqref{eq:generalform} and provided the
matrix $M_{\tau,\tau',\sigma}$, given by
\begin{equation}\label{eq:Mtts}
    M_{\tau,\tau',\sigma} = \begin{pmatrix}
    \frac{I}{\tau} & 0 & -K^T \\ 0 & \frac{I}{\tau'} & -{K'}^T \\ 
    -K & -K' & \frac{I}{\sigma}
    \end{pmatrix}
\end{equation}
is positive-definite. Before exploiting the estimate~\eqref{eq:ratePD}, let us express the conditions on
$\tau,\tau',\sigma$ which ensure that this is true. We need that for any $(\xi,\xi',\eta)\neq 0$,
\[
\frac{1}{\tau}\|\xi\|^2+\frac{1}{\tau'}\|\xi'\|^2+\frac{1}{\sigma}\|\eta^2\|
 > 2\sca{K\xi}{\eta}+2\sca{K'\xi'}{\eta}
\]
and obviously, this is the same as requiring that for any $a,a',b$ positive numbers,
\[
\frac{a^2}{\tau} + \frac{a'^2}{\tau'} + \frac{b^2}{\sigma} > 2(\|K\|a+2\|K'\|a')b.
\]
The worst $b$ in this inequality is $b=\sigma (\|K\|a+2\|K'\|a')$, then one checks easily
that the worse $a,a'$ are of the form $\bar a\|K\|\tau$, $\bar a\|K'\|\tau'$ respectively, so
that one should have for all $\bar a\neq 0$:
\[
\bar a^2 \left(\|K\|^2\tau + \|K'\|^2\tau'\right) > \sigma\bar a^2\left(\|K\|^2\tau + \|K'\|^2\tau'\right)^2,
\]
 yielding the condition
\begin{equation*}
    \sigma(\tau\|K\|^2+\tau'\|K'\|^2)<1.
\end{equation*}
We notice in addition that under such a condition, one also has 
\begin{equation*}
    M_{\tau,\tau',\sigma}\le 2\begin{pmatrix}
        \frac{I}{\tau} & &  \\  & \frac{I}{\tau'} &  \\  &  & \frac{I}{\sigma}
    \end{pmatrix}
\end{equation*}
which allows to simplify a bit the expression in the right-hand side of~\eqref{eq:ratePD} (at the expense
of a factor $2$ in front of the estimate).

We have not made use of the strong convexity up to now, and in particular, of~\eqref{eq:improveddescent}.
A quick look at the proof of Theorem~1 in~\cite{ChaPocMAPR16} shows that it will improve slightly
the latter condition, allowing to replace $\tau$ with the smaller effective step $\tau/(1+\tau\rho/4)$,
yielding the new condition
\begin{equation}\label{eq:newcondition}
    \sigma\left(\frac{\tau}{1+\tau\frac{\rho}{4}}\|K\|^2+\tau'\|K'\|^2\right)<1.
\end{equation}
This ensures now that~\eqref{eq:ratePD} holds with $M_{\tau,\tau',\sigma}$ replaced with

\begin{equation}
\label{eq:Mttsrho}
    M_{\tau,\tau',\sigma,\rho} = \begin{pmatrix}
    \left(\frac{1}{\tau}+\frac{\rho}{2}\right)I & 0 & -K^T  0 & \frac{I}{\tau'} & -{K'}^T \\ 
    -K & -K' & \frac{I}{\sigma}
    \end{pmatrix}
    \le 
    \begin{pmatrix}
    \left(\frac{2}{\tau}+\frac{3\rho}{4}\right)I & 0 & 0 
    \\ 0 & \frac{2}{\tau'}I & 0 \\ 
    0 & 0 & \frac{2}{\sigma}I
    \end{pmatrix}
\end{equation}
where the last inequality follows from~\eqref{eq:newcondition}.
Applied to problem~\eqref{eq:mainsaddle}, which is $\rho$-convex in $\mu$, we find that~\eqref{eq:newcondition} becomes the condition
\begin{equation}\label{eq:condtausig}
    \boxed{\sigma\left(\frac{\tau_\mu}{1+\tau_\mu\frac{\rho}{4}}\|Y\|^2
    +\tau\|X\|^2\right)<1.}
\end{equation}
When~\eqref{eq:condtausig} holds, then the ergodic iterates
(here we denote $W^n$, etc, the value of $W$ computed at the end of iteration $n$):
\begin{equation}\label{eq:ergodic}
\bar W^N = \frac{1}{N}\sum_{n=1}^N W^n, \  
{\bar\mu}^N = \frac{1}{N}\sum_{n=1}^N {\mu}^n, \  
\bar Z^N = \frac{1}{N}\sum_{n=1}^N Z^n.
\end{equation}
satisfy for all $W,\mu,Z$:
\begin{multline}
\label{eq:ratePD2}
    \mathcal L(\bar W^N,\bar \mu^N,Z) - \mathcal L(W,\mu,\bar Z^N) 
    \le 
    \frac{1}{N} \Big(\frac{3\rho}{8}\|\mu-\mu^0\|^2 +\frac{\|\mu-\mu^0\|^2}{\tau_\mu} +
    \frac{\|W-W^0\|^2}{\tau}+ \frac{\|Z-Z^0\|^2}{\sigma}\Big).
\end{multline}
Here the Lagrangian $\mathcal{L}$ is:
\[
\mathcal L(W,\mu,Z) = \sca{Z}{Y\mu-XW} + \lambda\|W\|_1 + \frac{\rho}{2}\|I-\mu\|^2.
\]
We denote $\mathcal E(W,\mu)= \sup_{Z}\mathcal L(W,\mu,Z)$ the primal energy (which appears
in~\eqref{L1}) and remark that in~\eqref{eq:mainsaddle}, $Z$ is bounded ($|Z_{i,j}|\le 1$ for all $i,j$) so that
$\|Z-Z^0\|^2\le 4mk$ in~\eqref{eq:ratePD2}. Hence, taking the supremum on $Z$ and
choosing for $(W,\mu)$ a primal solution (minimizer of $E$) $(W^*,\mu^*)$, we deduce:
\begin{multline}
\label{eq:rateprimal}
    \mathcal E(\bar W^N,\bar \mu^N) - \mathcal E(W^*,\mu^*) \le 
    \frac{1}{N} \left(\frac{4mk}{\sigma}+\Big(\frac{3\rho}{8}+\frac{1}{\tau_\mu}\Big)\|\mu^*-\mu^0\|^2+
    \frac{\|W^*-W^0\|^2}{\tau}\right).
    \end{multline}

In general, if one can compute reasonable estimates $\Delta_\mu$,
$\Delta_W$ for these quantities, one should take:
\begin{equation*}
\begin{split}
& \tau = \frac{\Delta_W}{2\sqrt{mk}\|X\|}, \ \tau_\mu =
\begin{cases}
\frac{1}{\frac{2\sqrt{mk}\|Y\|}{\Delta_\mu}-\frac{\rho}{4}} & \textup{ if }
\frac{8\sqrt{mk}\|Y\|}{\rho\Delta_\mu}>1\\
\tau_\mu>> 1 & \textup{ else,}
\end{cases}
\\
&
\sigma = \frac{1}{\frac{\tau_\mu}{1+\tau_\mu\frac{\rho}{4}}\|Y\|^2+\tau\|X\|^2}.
\end{split}
\end{equation*}
to obtain (considering here only the case $\rho$ small, that is when 
$\rho\Delta_\mu\le 8\sqrt{mk}\|Y\|$):
\begin{equation}
    \mathcal E(\bar W^N,\bar \mu^N) - \mathcal E(W^*,\mu^*) \le 
    \frac{\sqrt{mk} (5\Delta_\mu \|Y\| +4\Delta_W \|X\|)}{N},
\end{equation}
There is no clear way how to estimate \emph{a priori} the norm  $\|W^*-W^0\|$ in the Lagrangian approach.

\begin{rmrk}
Note that for the $\ell_1$  constrained problem~\eqref{constraint} $\Delta_W$ is bounded.
Since $\|W\|_1\le \eta$:  $\|W^*-W^0\|\le \|W^*-W^0\|_1\le 2\eta$, we use the estimate $\Delta_W\le 2\eta $. Using the initial value $\mu^0=I_k$,
$\Delta_\mu$ is also easily shown to be bounded (as $W$ is).
Empirically, we found that we can use the estimate $\Delta_\mu\lesssim\beta \|I_k \|_F= \beta \sqrt{k}$ where $\beta $ is a parameter to be tuned. Thus $\rho$ being small we have $\frac{8\sqrt{m}\|Y\|}{\rho\beta}>1$. Moreover, using $\|X\|=1$ ($X$ can be normalized),
we obtain the following reasonable choice of parameters:

\begin{equation} \label{3steps}
\begin{split}
& \tau = \frac{\Delta_W}{2\sqrt{mk}}\,, \quad \tau_\mu= 
\frac{\beta}{2\sqrt{m}\|Y\|-(1/4)\beta \rho}\,\\
& \sigma = \frac{1}{\frac{\tau_\mu}{1+\tau_\mu\frac{\rho}{4}}\|Y\|^2+\tau} \cdot
\end{split}
\end{equation}

\end{rmrk}

\noi In the case of Problem~\eqref{L2saddle} (Sec.~\ref{fro}), $Z$ is also
bounded but then, one has simply $\|Z^*-Z^0\|^2\le 4$, hence~\eqref{eq:rateprimal} must
be replaced with
\begin{equation}
\label{eq:rateprimalfro}
    \mathcal E(\bar W^N,\mu^N) - \mathcal E(W^*,\mu^*) \le 
    \frac{1}{N} \left(\frac{4}{\sigma}+\Big(\frac{3\rho}{8}+\frac{1}{\tau_\mu}\Big)\|\mu^*-\mu^0\|^2+    \frac{\|W^*-W^0\|^2}{\tau}\right).
\end{equation}
(Obviously, now, the energy $\mathcal E$ is the primal energy in~\eqref{L2}.)
The same analysis as before remains valid, but now with $mk$ replaced with $1$.

\subsection{Convergence with over-relaxation }\label{s4.2}
For the over-relaxed variant (Algorithm~\ref{L1-SR}), the adaption is a little bit more
complicated, and one does not benefit much from taking into account the
partial strong convexity.
One approach is to rewrite the improved descent rule~\eqref{eq:improveddescent}
as follows:
\begin{multline}
\label{eq:improveddescent2}
f(\hat x) \le f(x) + 
 \frac{1+\tau\rho/2}{2\tau} \big( \|x-\bar x\|^2-\|x-\hat x\|^2
 \\-\|\hat x-\bar x\|^2 \big) + \frac{\rho}{8}\|\hat x-\bar x\|^2 \\ =
 f(x) + 
 \frac{1}{2\tilde \tau} \sca{x-\hat x}{\hat x-\bar x} + \frac{\rho}{8}\|\hat x-\bar x\|^2
\end{multline}
where $\tilde\tau = \tau/(1+\tau\rho/2)$ is an effective time-step.
As a result, we observe that the first (primal) update in~\eqref{eq:primaldual} yields
the same rule as an explicit-implicit primal update of a nonsmooth$+$smooth functions
with effective step $\tilde\tau$ and Lipschitz constant $\rho/4$, \textit{cf}
Eq.~(9) in~\cite{ChaPocMAPR16}.
Hence, the analysis of these authors 
(see Sec.~4.1 in the above reference) can be reproduced almost identically and will
 yield for the over-relaxed algorithm~\eqref{L1-SR} similar convergence rates,
 cf.~\eqref{eq:ratePD}-\eqref{eq:rateprimal}, now,
 with the factor $1/N$ replaced with $1/((1+\gamma)N)$. It requires that the matrix
 \begin{equation}\label{eq:Mt}
    \tilde M = \begin{pmatrix}
    (\frac{1}{\tilde\tau}- \frac{\rho/4}{1-\gamma}) I & 0 & -K^T \\ 0 & \frac{I}{\tau'} & -{K'}^T \\ -K' & -K & \frac{I}{\sigma}
    \end{pmatrix}
\end{equation}
be positive definite. 
 Observe however that the estimates hold for the ergodic averages (cf~\eqref{eq:ergodic})
 of the variables obtained at the end of Step~\ref{statefinal1} of Algorithm~\ref{L1-SR}
 and Step~\ref{statefinal2} of Algorithm~\ref{PD-L2-OR},
 rather than for the over-relaxed variables
 (which could not even be feasible).
 We derive that for this method, condition~\eqref{eq:condtausig} should
 be replaced with
 \begin{equation}\label{eq:condOR}
     \sigma\left(\frac{\tau_\mu}{1+\frac{\tau_\mu\rho}{4}\frac{1-2\gamma}{1-\gamma}}\|Y\|^2+\tau\|X\|^2\right)<1,
 \end{equation}
 at least if $\gamma<1/2$.
 
 As seen, for $\gamma\ge 1/2$, the condition obtained for $\rho=0$ 
 is better (hence, the
 partial strong convexity does not seem to yield any reasonable improvement
 for this algorithm). It simply reads
\begin{equation}\label{eq:condbase}
     \sigma\left(\tau_\mu\|Y\|^2+\tau\|X\|^2\right)<1,
\end{equation}
and one gets the estimate from~\cite{ChaPocMAPR16} (Eq.~(24), further simplified thanks
to~\eqref{eq:condOR}):
\begin{equation}
\label{eq:rateprimalOR}
    \mathcal E(\bar W^N,\bar \mu^N) - \mathcal E(W^*,\mu^*) \le 
    \frac{1}{(1+\gamma)N} \left(\frac{4mk}{\sigma}+\frac{\|\mu^*-\mu^0\|^2}{\tau_\mu}+
    \frac{\|W^*-W^0\|^2}{\tau}\right).
\end{equation}

\section{Derivation of the min-max iteration} \label{sA}
As explained in Section~\ref{s4.1}, we consider the following general min-max problem:
\begin{equation} \label{eq61}
  \min_{(x,x')} \max_y f(x)+g(x') + \sca{Kx+K'x'}{y} - h^*(y)
\end{equation}
for convex functions $f$, $g$, $h$, and linear operators $K$, $K'$. Note that, since
$h^*$ is the convex conjugate of $h$, for any fixed $x'$ one has
\[ \max_y \sca{Kx+K'x'}{y} - h^*(y) = h^{**}(Kx+K'x') = h(Kx+K'x'), \]
so that the problem can also be rewritten as
\[ \min_{(x,x')} f(x)+g(x') + h(Kx+K'x'). \]
In our situation, we dualize the computation of the $\ell_1$ norm containing the
linear terms according to
\[ \|Y\mu-XW\|_1 = \max_{\|Z\|_\iy \leq 1} \sca{Z}{Y\mu-XW}. \]
As a result, the original minimization
\[ \min_{(W,\mu)} \|Y\mu-XW\|_1 + \lambda\|W\|_1+\frac{\rho}{2}\|\mu-I\|_F^2 \]
is changed into the min-max problem
\[ \min_{(W,\mu)} \max_Z \lambda\|W\|_1+\frac{\rho}{2}\|\mu-I\|_F^2
 + \sca{Z}{Y\mu-XW} - \delta_{B_\iy}(Z) \]
where $\delta_{B_\iy}$ denotes the indicator function of the $\ell_\iy$ unit ball.
This problem fits in our general min-max framework by setting $(x,x'):=(W,\mu)$,
$y:=Z$, together with
\[ f(W) := \lambda\|W\|_1,\quad
   g(\mu) := \frac{\rho}{2}\|\mu-I\|_F^2, \]
\[ h^*(Z) := \delta_{B_\iy}(Z), \]
and $Kx+K'x' = -XW+Y\mu$. (Note that, as the conjugate of a norm is the indicatrix
of the unit ball of the dual norm, one indeed has $h(z)=\|z\|_1$.) Similarly, when replacing the $\ell_1$ norm with the Huber function for the loss term, one has (in Lagrangian form)
\[ \min_{(W,\mu)} h_\delta(Y\mu-XW)+\lambda\|W\|_1
   +\frac{\rho}{2}\|\mu-I\|_F^2 \]
which is dualized according to (\ref{eq61}) where $f$ and $g$ are defined as before, and where one takes $h_\delta$ instead of the $\ell_1$-norm for $h$. Using the fact that
\[ h_\delta^*(s)=\delta s^2/2+\delta_{[-1,1]}(s) \]
and vectorizing the computation, one obtains
\[ \min_{(W,\mu)} \max_{\|Z\|_\iy \leq 1} \sca{Z}{Y\mu-XW}+\lambda\|W\|_1
   +\frac{\rho}{2}\|\mu-I\|_F^2 - \frac{\delta}{2}\|Z\|_F^2, \]
from where one retrieves (\ref{eq:mainsaddle}).

\bibliography{arXiv}
\bibliographystyle{abbrvnat}
\end{document}